\documentclass[lettersize,journal]{IEEEtran}
\usepackage{amsmath,amsfonts}
\usepackage{algorithmic}
\usepackage{algorithm}
\usepackage{array}
\usepackage[caption=false,font=normalsize,labelfont=sf,textfont=sf]{subfig}
\usepackage{textcomp}
\usepackage{stfloats}
\usepackage{url}
\usepackage{verbatim}
\usepackage{graphicx}
\usepackage{cite}
\usepackage{multirow}
\usepackage{color}
\usepackage{bm}
\usepackage{bbding}
\usepackage{pifont}
\usepackage{makecell}
\usepackage{colortbl}
\usepackage[table]{xcolor}
\usepackage{microtype}
\usepackage[colorlinks,linkcolor=red,anchorcolor=blue,citecolor=green]{hyperref}
\usepackage{booktabs}
\usepackage{wrapfig}
\newfloat{figtab}{htb}{fgtb}
\usepackage{subfiles}
\newcommand{\cmark}{\ding{51}}%
\newcommand{\xmark}{\ding{55}}%
\definecolor{cred}{HTML}{FF6B6B}
\definecolor{cyellow}{HTML}{FEC260}
\definecolor{cgreen}{HTML}{70AD47}
\definecolor{cblue}{HTML}{4D96FF}
\definecolor{cpurple}{HTML}{2A0944}
\definecolor{ggray}{RGB}{127,127,127}
\definecolor{aliceblue}{rgb}{0.94, 0.97, 1.0}
\definecolor{bblue}{RGB}{0,30,95}
\definecolor{rred}{RGB}{190,0,0}
\definecolor{mygray}{gray}{.9}
\definecolor{ggray}{RGB}{127,127,127}
\definecolor{sblue}{RGB}{0,173,206}
\definecolor{ppink}{RGB}{240,46,142}
\newcommand{\sssection}[1]{\noindent\textbf{#1}}
\def\eg{\emph{e.g.}} 
\def\ie{\emph{i.e.}} 
\def\cf{\emph{c.f.}} 
\def\etc{\emph{etc.}} 
\newcommand{\pub}[1]{\color{gray}{\tiny{[{#1}]}}}

\makeatletter
\newcommand{\thickhline}{%
    \noalign {\ifnum 0=`}\fi \hrule height 1pt
    \futurelet \reserved@a \@xhline
}
\makeatother

\begin{document}

\title{MVP-Shot: Multi-Velocity Progressive-Alignment Framework for Few-Shot Action Recognition}

\author{IEEE Publication Technology,~\IEEEmembership{Staff,~IEEE,}
\thanks{This paper was produced by the IEEE Publication Technology Group. They are in Piscataway, NJ.}
\thanks{Manuscript received April 19, 2021; revised August 16, 2021.}}

\author{Hongyu~Qu, Rui~Yan, Xiangbo~Shu, Hailiang~Gao, Peng~Huang, and Guo-Sen~Xie 
\thanks{\textit{H. Qu, X. Shu, H. Gao, P. Huang and G.-S. Xie are with the School of Computer Science and Engineering, Nanjing University of Science and Technology, Nanjing 210094, China. E-mail: \{quhongyu, shuxb, gaohailiang, penghuang\}@njust.edu.cn, gsxiehm@gmail.com. (Corresponding author: Xiangbo Shu)}}
\thanks{\textit{R. Yan is with the Department of Computer Science and Technology, Nanjing University, Nanjing 210023, China. E-mail: ruiyan@nju.edu.cn.}}
}


\maketitle

\begin{abstract}
Recent few-shot action recognition (FSAR) methods typically perform semantic matching on learned discriminative features to achieve promising performance. However, most FSAR methods focus on single-scale (\eg, frame-level, segment-level, \etc) feature alignment, which ignores that human actions with the same semantic may appear at different velocities. To this end, we develop a novel Multi-Velocity Progressive-alignment (MVP-Shot) framework to progressively learn and align semantic-related action features at multi-velocity levels. Concretely, a Multi-Velocity Feature Alignment (MVFA) module is designed to measure the similarity between features from support and query videos with different velocity scales and then merge all similarity scores in a residual fashion. To avoid the multiple velocity features deviating from the underlying motion semantic, our proposed Progressive Semantic-Tailored Interaction (PSTI) module injects velocity-tailored text information into the video feature via feature interaction on channel and temporal domains at different velocities. The above two modules compensate for each other to make more accurate query sample predictions under the few-shot settings. Experimental results show our method outperforms current state-of-the-art methods on multiple standard few-shot benchmarks (\ie, HMDB51, UCF101, Kinetics, and SSv2-small).
\end{abstract}
\begin{figure}[!t]
    \centering
    \includegraphics[width=0.48\textwidth]{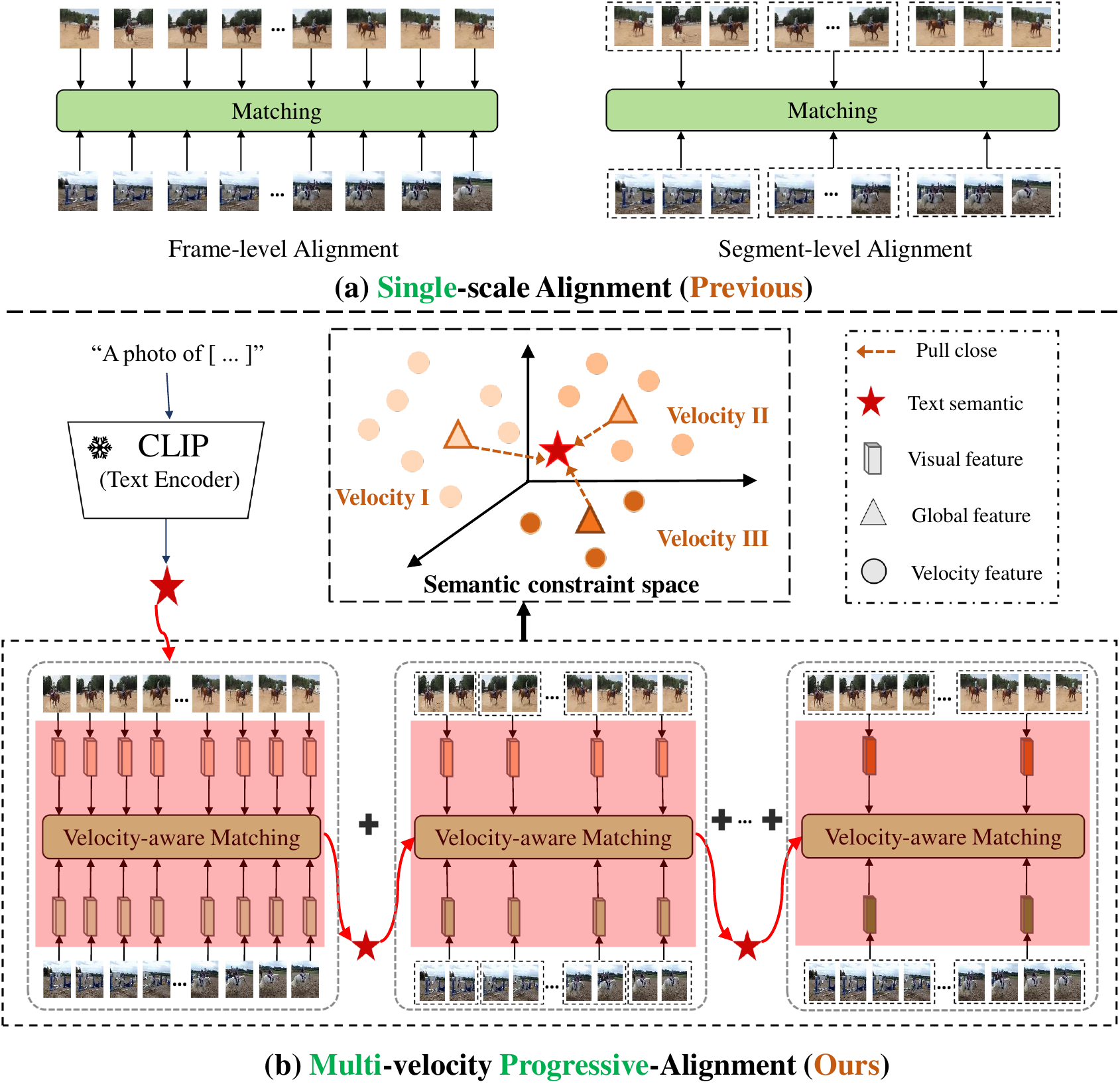}
    \setlength{\abovecaptionskip}{0.1cm} 
    \caption{The main idea of this work. (a): The previous methods focus on {\bf single-velocity feature alignment}, such as frame-level and segment-level. (b): In our work, we capture multi-velocity feature representation and perform {\bf multi-velocity feature alignment} to deal with action instances with diverse speeds.}
    \label{fig1}
    \vspace{-0.7cm}
\end{figure}
\section{Introduction}
Due to the increasing demand for video analysis in real-world scenarios, significant progress has been made in action recognition~\cite{slowfast,tsm,qing2023mar,wang2023learning,huang2023semantic} recently. Despite this, this success still relies heavily on extensive manually-labeled samples, and collecting these samples is expensive and labor-intensive. This significant bottleneck severely constrains the broader applicability of this task. To alleviate the data-hunger issue, few-shot action recognition (FSAR) is a promising direction and has gained widespread attention~\cite{zhu2018compound,wang2022hybrid,huang2022compound,boosting,shi2024commonsense} in recent years. The objective of FSAR is to identify unseen action categories with only a few labeled video examples.
 

Current mainstream FSAR approaches are built upon the metric-based meta-learning paradigm, which learns discriminative class prototype representation and predicts action category labels based on learned distance metrics.  Most metric-based methods perform \textbf{frame-level alignment}~\cite{cao2020few,wang2022hybrid,ta2n,thatipelli2022spatio,boosting} on learned discriminative frame-level features, which measure the similarity between different videos via diverse matching strategies~\cite{cao2020few,wang2022hybrid,perrett2021temporal}. In addition, some methods perform \textbf{segment-level alignment}~\cite{bishaytarn,ben2021taen,zhang2020few} on obtained segment-level features. Typically, TARN~\cite{bishaytarn} first uniformly samples segments from videos, and performs segment-level alignment based on segment features extracted by 3D-CNN. Although existing FSAR methods have achieved remarkable performance, these methods still mainly focus on single-scale feature alignment (\ie, frame-level alignment and segment-level alignment),  as shown in Fig.~\ref{fig1}{\color{red}(a)}, which is insufficient to reflect the diversity of action speeds. Therefore, these single-scale approaches often result in inaccurate matching when dealing with action instances with diverse speeds. In real-world scenarios, different individuals may perform the same action class at different velocities. Take the action ``ride horse" as an example, different persons act at different velocities. Specifically from a visual perspective, certain horses may be perceived sprinting at a rapid pace, while others may be observed moving at a more leisurely walk. 

\textbf{In our paper, we focus on multi-velocity feature alignment in FSAR}, where multi-velocity features are utilized to capture pair-wise semantic relevance at different velocity scales. With multi-velocity feature alignment, we can achieve more accurate matching to some extent in the collaboration of multiple-velocity features. However, with only limited support videos, the learned multi-velocity action features are not up to adhere to the underlying motion semantic without extra semantic constraints. Actually, the human perception system possesses a distinctive visual mechanism known as cognitive penetrability~\cite{maier2019no}. This mechanism employs prior linguistic knowledge to adjust visual perception processing towards semantic-relevant features. Hence, we argue that it is necessary to leverage text information to avoid multi-velocity features deviating from the underlying motion semantic for the rare support samples.

 In light of the above insights, we propose a novel \textbf{Multi-Velocity Progressive-alignment (MVP-Shot)} framework to learn and align multi-velocity action features in a progressive manner (Fig.~\ref{fig1}{\color{red}(b)}). More specifically, we present a novel \textbf{Progressive Semantic-Tailored Interaction (PSTI)} module to capture multiple action features from different velocity scales, in which we recurrently inject velocity-tailored text information at different velocities to enhance velocity-aware semantic features. Thus rich semantic prior knowledge is fully utilized to avoid multiple velocity features deviating from the original motion semantic. To cooperate with the PSTI module,  we further design a \textbf{Multi-Velocity Feature Alignment (MVFA)} module to capture pair-wise semantic similarity scores at different velocity scales, and then merge all similarity scores in a residual fashion to facilitate coherent velocity-aware matching between different videos. In this way, our proposed MVP-shot combines the above two modules to compensate for each other, which enables more robust and more accurate few-shot feature matching. 
 
  We conduct experiments on four gold-standard datasets (\ie, HMDB51~\cite{kuehne2011hmdb}, UCF101~\cite{soomro2012ucf101}, Kinetics~\cite{carreira2017quo}, and SSv2-small~\cite{goyal2017something}) to comprehensively evaluate the effectiveness of our MVP-shot. We empirically prove that our MVP-shot surpasses all existing state-of-the-arts and yields solid performance gains,~\eg, $3.2\%$ accuracy gains on HMDB51 dataset on the $5$-way $1$-shot setting. Furthermore, we perform thorough ablation studies to dissect each component, both quantitatively and qualitatively. 

 In summary, the main contributions of this paper are as follows:
\begin{itemize}
    \item We propose a novel MVP-Shot framework to conduct Few-Shot Action Recognition (FSAR) matching by learning and aligning reliable multi-velocity action features in a progressive way.
    \item  {\bf To learn multi-velocity action features} reliably, we design a new Progressive Semantic-Tailored Interaction (PSTI) module that recurrently injects velocity-tailored text information into video features via feature interaction on different domains. 
    \item  {\bf To align multi-velocity action features} covering different temporal dynamics, we present a new Multi-Velocity Feature Alignment (MVFA) module to measure the similarity of action features from different velocities in a residual fashion. To the best of our knowledge, this is the pioneering effort to exploit multi-velocity matching for FSAR.
  \end{itemize} 

We organize the rest of this paper as follows: Section~\S\ref{related work} describes several lines of related works regarding few-shot image classification, few-shot action recognition, and transferring CLIP models for video recognition. Section~\S\ref{methodology} presents the proposed MVP-shot in detail for FSAR. Section~\S\ref{experiments} reports comparison results and analysis of extensive experiments. Finally, the conclusion is provided in Section~\S\ref{conclusion}.
\section{Related Work}
\label{related work}
In this section, we introduce several lines of research closely related to our MVP-shot, regarding few-shot image classification (\S\ref{21}), few-shot action recognition (\S\ref{22}), and transferring CLIP models for video recognition (\S\ref{23}).
\subsection{Few-Shot Image Classification}
\label{21}
The few-shot learning (FSL) task aims to recognize unseen classes with limited labeled samples. Mainstream FSL methods are roughly categorized into three groups: \textbf{i)} \textit{Augmentation-based} methods~\cite{chen2019image,li2020adversarial,ye2021learning} usually learn to increase the number of  features or training samples to augment model training and improve few-shot model's generalization ability, mainly including hallucinating additional training examples ~\cite{chen2019image} and adversarial generation ~\cite{li2020adversarial}; \textbf{ii)} \textit{Optimization-based} methods~\cite{finn2017model,jamal2019task,rajeswaran2019meta,rusu2018meta} learn an effective model initialization state or develop techniques to properly update model parameters, enabling rapid adaptation to new tasks with minimal fine-tuning steps for enhanced generalization.Typical methods include MAML~\cite{finn2017model} and other variants~\cite{jamal2019task,rajeswaran2019meta,rusu2018meta}; \textbf{iii)} \textit{Metric-based} methods~\cite{wang2019simpleshot,ye2020few,chen2023semantic,han2023textual,peng2023sgva,zhong2022graph,cheng2023graph} leverage visual encoders to map images into a common embedding space and predict query (\ie, test sample) labels with respect to different distance metrics, such as Euclidean distance~\cite{yoon2019tapnet,wang2019simpleshot} and cosine similarity~\cite{qiao2018few}. For example, Prototypical Network~\cite{snell2017prototypical} measures the similarity of test samples and the class prototypes of the support set by Euclidean distance metric and makes few-shot predictions based on obtained similarity scores. Our work belongs to the metric-based methods and tries to solve challenging few-shot action recognition (FSAR), which aims to identify unlabeled videos with complex spatiotemporal structures using only a few video examples. Our work focuses on making full use of temporal information to obtain and align semantic-related action features at multi-velocity levels.

\begin{figure*}[!t]
    \centering
    \includegraphics[width=0.99\textwidth]{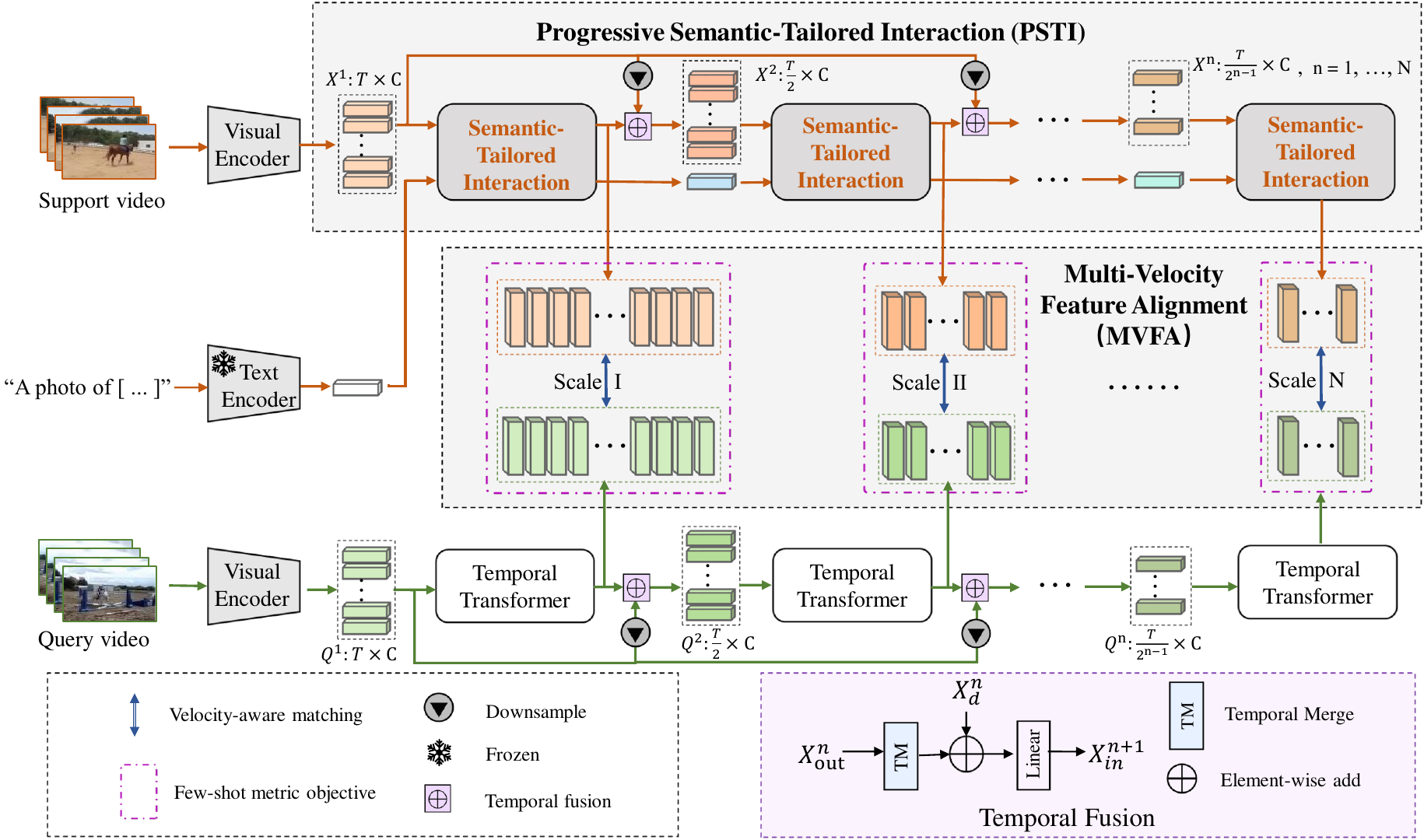}
    \setlength{\abovecaptionskip}{0.2cm} 
    \caption{The overview of the proposed Multi-Velocity Progressive-Alignment (MVP-shot) Framework. The support and query videos are first fed into a visual encoder to extract frame-level features. Then, based on frame features, we apply a Progressive Semantic-Tailored Interaction (PSTI) module to capture multi-velocity action features. Accordingly, we employ a Multi-Velocity Feature Alignment (MVFA) module to measure pair-wise semantic similarity at different velocity scales. For clarity, the figure does not depict other support videos in the few-shot task.}
    \label{fig2}
    \vspace{-0.05in}
\end{figure*}
\subsection{Few-Shot Action Recognition}
\label{22}
Few-shot action recognition (FSAR) learns discriminative class prototype representation and categorizes unlabeled query video samples with a few labeled samples per class (\ie, support samples) by performing prototype-query matching based on distance metrics. Different from images, video has temporal diversity and richer spatio-temporal contextual information. Therefore, directly applying few-shot image classification methods to few-shot video recognition tasks is not appropriate. The existing few-shot action recognition methods~\cite{wang2022hybrid,liu2023lite,wu2022motion,xing2023revisiting} mainly follow a metric-based learning paradigm to exploit the temporal cues. As the primary step, some approaches~\cite{tang2023m3net,wu2022motion} aim to learn robust feature representation for few-shot action recognition. STRM~\cite{thatipelli2022spatio} performs spatial feature interaction in each frame and temporal feature interaction by attention mechanisms in a cascaded manner, so as to learn discriminative feature representations. SloshNet~\cite{xing2023revisiting} designs an architecture search module for feature fusion, aiming to automatically discover the optimal integration of spatial features from different levels. ITANet~\cite{zhang2021learning} introduces joint spatiotemporal modeling techniques to achieve implicit temporal alignment and matching. HyRSM~\cite{wang2022hybrid} makes full use of all videos in a meta-learning task to learn task-specific patterns by hybrid relation module. Other methods attempt to design robust class prototype matching~\cite{cao2020few,cao2022searching,li2024hierarchical} strategies. Among them,  OTAM~\cite{cao2020few} maintains the frame order in video and aligns the two video sequences using an improved dynamic time-warping algorithm. TARN~\cite{bishaytarn} utilizes an attention mechanism between segment visual features to achieve temporal alignment. TRX~\cite{perrett2021temporal} establishes query-specific video matching by matching sub-sequences between different videos. In a word, the aforementioned methods focus on segment-level alignment or frame-level alignment solely. However, these single-scale approaches do not handle action instances with diverse speeds effectively. Unlike these previous methods, our work combines multi-velocity feature representation and multi-velocity feature alignment to compensate for each other, which enables more robust and more accurate few-shot matching.

\subsection{Transferring CLIP Models for Video Recognition}
\label{23}
With deep learning advancements~\cite{tang2016tri,shu2015weakly,shu2019hierarchical,shu2021spatiotemporal}, significant progress has been made in the field of action recognition~\cite{yan2023progressive,shu2022multi,tang2019coherence,huang2024appearance} in recent years. Recently, image-language pretraining\cite{luo2021coco,radford2021learning} has achieved remarkable progress due to its powerful visual representations aligned with rich linguistic semantics. CLIP\cite{radford2021learning} is one of the most representative works, which achieves impressive “zero-shot” transferability and generalization capacity. CLIP is a vision-language transformer capable of jointly understanding both images and text by contrastive learning. Various downstream tasks begin to widely adopt these pre-trained models, showcasing impressive performance across the board, including image classification\cite{zhou2022conditional,chen2023semantic} and video understanding\cite{xu2021videoclip,ju2022prompting}. Inspired by these successes, CLIP-FSAR\cite{wang2023clip} firstly makes use of CLIP to integrate auxiliary semantic information (provided by class names) and visual features for FSAR. However, CLIP models are trained on large-scale image pairs, and different from images, videos contain richer spatiotemporal contextual information. Current methods only propose an additional temporal module to extend image pre-trained models, which are unable to take full advantage of temporal information in videos. To perform temporal modeling effectively, we propose a Semantic-Tailored Interaction module including two complementary sub-modules to inject semantic information provided by class names into the visual features, which performs feature interaction on the temporal and the channel dimensions. The Semantic-Tailored Interaction module fully utilizes semantic information provided by class names to obtain more discriminative video features and enhance video class prototype representation for few-shot learning.
\section{Method}
\label{methodology}
In this section, we first introduce the problem statement of the few-shot action recognition task (\S\ref{problem}). Then, as shown in Fig.~\ref{fig2}, an architecture overview of our proposed MVP-shot is presented (\S\ref{arch}). Finally, we describe in detail each component of our proposed MVP-shot (\S\ref{progressive} and \S\ref{multi-velocity}).
\subsection{Problem Statement}
\label{problem}
The goal of few-shot action recognition is to learn a model that has a strong generalization capability to identify unseen categories given a few (\eg, one or five) labeled samples per class. In the few-shot learning setting, the whole dataset is divided into a training set $\mathcal{D}_{base}= \{ (v_{i}, y_{i}), y_{i} \in \mathcal{C}_{base} \} $ and a testing set $\mathcal{D}_{novel}=\{ (v_{i},y_{i}), y_{i} \in \mathcal{C}_{novel} \} $, where $y_{i}$ denotes the action label of video $v_{i}$. Note that the categories of $\mathcal{D}_{base}$ and $\mathcal{D}_{novel}$  are sample-wise non-overlapping, \ie, $\mathcal{C}_{base}\cap\mathcal{C}_{novel}=\emptyset$. Under the few-shot setting, we follow the episodic training paradigm~\cite{vinyals2016matching} to train a meta-learning model for FSAR. In each episodic task (\ie, a standard \textit{N}-way \textit{K}-shot task), there are two sets (\ie,  query set \textit{Q} and support set \textit{S}) sampling \textit{N} different action categories from training set. The support set consists of \textit{K} support videos per category, which typically range from $1$ to $5$, and the query set consists of a fraction of the rest videos from \textit{N} action categories. The goal is to correctly classify the query sample into one of the \textit{N} categories in the support set. During the inference phase, we randomly sample multiple episodic tasks from $\mathcal{D}_{novel}$ to evaluate the model's generalization capabilities.

\subsection{Architecture Overview}
\label{arch}
We illustrate the overall architecture of our proposed MVP-shot in Figure~\ref{fig2}. In a standard \textit{N}-way \textit{K}-shot episodic task, the support set $S=\{s_{1}, s_{2},...,s_{N\times K} \}$ consists of $N$ action categories and $K$ videos per category, where $s_{i}\in \mathbb{R} ^{T\times 3 \times H\times W}$  and $T$ is the number of sparsely sampled video frames. For the sake of clarity, we present our framework using $N$-way $1$-shot task, and a single video $q$ is included in the query set $Q$. The visual encoder of CLIP first processes video inputs to generate query video feature $f_{q}$ and the support video features $F_{S}=\{f_{s_1},f_{s_2},...,f_{s_N} \}$, where $f_{i}=\{f^{1}_{i},f^{2}_{i},...,f^{T}_{i}  \}, f^{j}_{i}\in \mathbb{R} ^{C}$ and $C$ denotes the channel number. In addition, the corresponding category names are input into the text encoder of CLIP to obtain textual features. Then we propose a Progressive Semantic-Tailored Interaction (PSTI) module to obtain semantic-related action features from different velocities. With such multi-velocity feature representation, we utilize a Multi-Velocity Feature Alignment (MVFA) module to measure the similarity of the specific semantic information at different temporal scales and produce the final predicted probabilities after a fusion operation. In our proposed MVP-shot, visual features from different velocity scales are complementary, and they work together to enable more robust and more accurate few-shot feature matching.

\subsection{Progressive Semantic-Tailored Interaction}
\label{progressive}
For some challenging action categories with diverse speeds, only utilizing single-velocity temporal features may lead to less satisfactory video classification results.  Thus, we design a novel Progressive Semantic-Tailor Interaction (PSTI) module to capture temporal features from different velocity scales, where each temporal feature captures unique perspectives of temporal representations. In FSAR, with limited support videos, the learned velocity features tend to deviate from the underlying motion semantic without semantic information constraints. We argue that textual and visual information compensates for each other, and using linguistic prior knowledge promotes the visual features to 
focus more on semantic-related features. To this end, PSTI consists of multiple Semantic-Tailor Interaction blocks. Each block progressively utilizes the output text information of the last stage to enhance current semantic-related velocity features, and then each stage outputs corresponding velocity action features for velocity-aware matching. 
\begin{figure}[!t]
      \centering
    \includegraphics[width=0.37\textwidth]{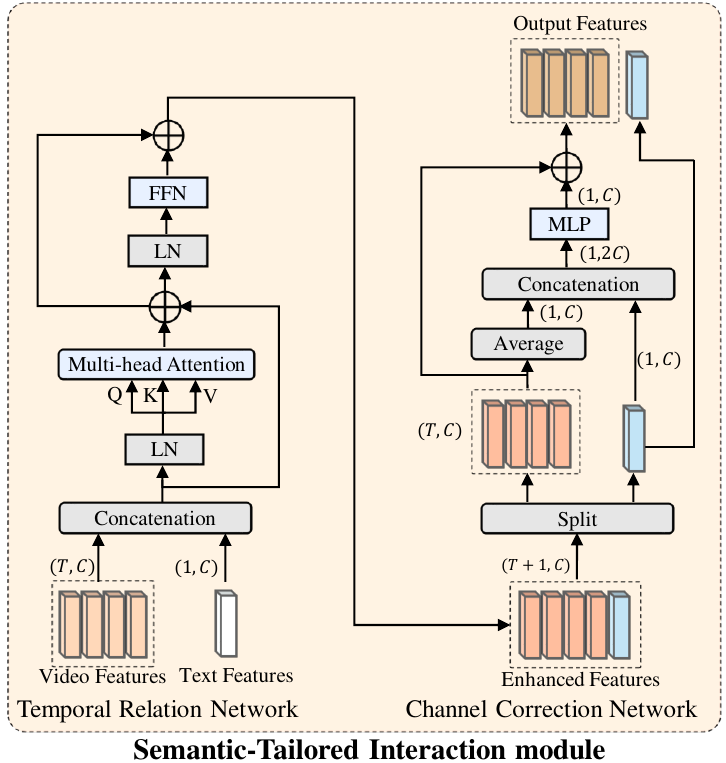}
    \setlength{\abovecaptionskip}{0.2cm}
    \caption{Illustration of Semantic-Tailored Interaction (STI) module. STI consists of Temporal Relation Network and Channel Correction Network, which perform feature interaction in the temporal and channel domains, respectively. }
    \label{fig3}
    \setlength{\belowcaptionskip}{-0.5cm}
\end{figure}

As shown in Figure~\ref{fig3}, the Semantic-Tailor Interaction block (STI) comprises two sub-networks: Temporal Relation Network and Channel Correction Network. These two networks compensate each other by injecting semantic information provided by class names into visual features, which perform feature interaction on the temporal and the channel
dimensions, respectively. Take an STI block as an example. In the STI block that processes information at the velocity scale $n$, the input visual features and textual features are $\bm{X}_{\mathrm{in}}^{n}\in {\mathbb{R}}^{T\times C}$ and $\bm{q}_{\mathrm{in}}^{n}\in {\mathbb{R}}^{1\times C}$, respectively. 

\noindent \textit{(a) Temporal Relation Network:} Inspired by the prompt methods in NLP~\cite{brown2020language}, we concatenate the visual features with the corresponding text feature,  and feed them together into the Temporal Relation Network to obtain the enhanced features. Concretely, given the text feature $\bm{q}_{\mathrm{in}}^{n}\in {\mathbb{R}}^{1\times C}$  and the visual features $\bm{X}_{\mathrm{in}}^{n}\in {\mathbb{R}}^{T\times C}$ at the velocity scale $n$, we obtain new  sequence features $\hat{\bm{X}^{n}}$ by concatenating the text features and the visual features.

Then we feed the extended sequence features $\hat{\bm{X}^{n}}$ into the remaining transformer layers, which consist of Multi-Headed Self-Attention (MSA), and MLP layer to allow the interaction between the semantic token and the visual tokens along the temporal dimension. Concretely, we compute the interim interaction output $\dot{\bm{X}^{n}}$ as 
\begin{equation}
      \dot{\bm{X}^{n}} = \hat{\bm{X}^{n}}+\mathrm{MSA}(\hat{\bm{X}^{n}}),     
\end{equation}
\begin{equation}            
    \mathrm{MSA}^{h}=\mathrm{Softmax}(\frac{{\bm{F}_{\mathrm{q}}^{h}}{{\bm{F}_{\mathrm{k}}^{h}}^\top}}{\sqrt{d}})\bm{F}_{\mathrm{v}}^{h},
\end{equation}
where $\mathrm{MSA}^{h}$ is a single attention head $h$ in MSA, $\bm{F}_{\mathrm{q}}^{h} =\mathrm{LN}(\hat{\bm{X}^{n}})\bm{W}_{\mathrm{q}}^{h} $ is the query, $\bm{F}_{\mathrm{k}}^{h} =\mathrm{LN}(\hat{\bm{X}^{n}})\bm{W}_{\mathrm{k}}^{h} $ is the key, and $\bm{F}_{\mathrm{v}}^{h} =\mathrm{LN}(\hat{\bm{X}^{n}})\bm{W}_{\mathrm{v}}^{h} $ is the value. We concatenate all the attention heads and perform linear projection to obtain the final output $\dot{X^{n}}$. Then, as shown in Fig. 3, we further apply FFN to obtain enhanced feature $\bar{\bm{X}^{n}}$, given by:
\begin{equation}
      \bar{\bm{X}^{n}} = \dot{\bm{X}^{n}}+\mathrm{FFN}(\mathrm{LN}(\dot{\bm{X}^{n}})),   
\end{equation}
where FFN consists of two linear projections separated by a GELU activation function.

\noindent \textit{(b) Channel Correction Network:} In addition to fully accounting for temporal interaction between the visual features and the text semantic, we believe that taking advantage of text semantic information to perform channel interaction is conducive to multi-velocity feature construction. Thus, we propose a Channel Correction Network that utilizes the channel interaction to correct feature channels and enhance semantic-related velocity features. Specifically, given enhanced visual features $\bar{\bm{X}^{n}}=[\bm{x}_{1},\bm{x}_{2},...,\bm{x}_{T}]$, a global visual context vector $\bar{\bm{x}_{\mathrm{g}}^{n}}$ is firstly obtained by averaging all visual tokens:
\begin{equation}
         \bar{x_{g}^{n}} = \frac{1}{T}\sum_{t=1}^{T}x_{t}.
\end{equation}
Then, we concatenate the global visual context vector with enhanced text feature $\bar{\bm{q}^{n}}$. Specifically, we employ a two-layer MLP module to learn relation-aware channel features and obtain a modulating global vector $\bm{p}_{\mathrm{g}}^{n} \in {\mathbb{R}}^{1\times C}$:
\begin{equation}
         \bm{p}_{\mathrm{g}}^{n} = \mathrm{MLP}([\bar{\bm{x}_{\mathrm{g}}^{n}};\bar{\bm{q}^{n}}]),
\end{equation}
\begin{equation}
         \mathrm{MLP}(\cdot) =\mathrm{FC}(\mathrm{ReLU}(\mathrm{FC}(\cdot))), 
\end{equation}
where $\mathrm{MLP}(\cdot)$ denotes the MLP layer, which comprises two fully connected layers and a ReLU activation layer. Finally, we add the modulated global vector to all visual tokens so that visual features pay more attention to semantic information. The final modulated visual features $\bm{X}_{\mathrm{out}}^{n} \in {\mathbb{R}}^{T\times C}$ can be written as:
\begin{equation}
          \bm{X}_{\mathrm{out}}^{n} = [\bm{x}_{t}+\bm{p}_{\mathrm{g}}^{n},],  t=1,2,...,T.
\end{equation}

In addition, we design progressive connection operation to connect the visual outputs of the STI block at different velocity scales and pool features from the raw video input ${\bm{X}_{\mathrm{in}}^{n}}$, so as to preserve high-level scene information. Given the interaction feature output $\bm{X}_{\mathrm{out}}^{n}\in {\mathbb{R}}^{T\times C}$, the progressive connection uses temporal fusion to provide $\bm{X}_{\mathrm{in}}^{n+1} \in {\mathbb{R}}^{\frac{T}{2}\times C}$ for the next block, as follows,
\begin{equation}
          \bm{X}_{\mathrm{in}}^{n+1} = \mathrm{Linear}(\mathrm{TM}(\bm{X}_{\mathrm{out}}^{n})+\bm{X}_{\mathrm{d}}^{n}),
\end{equation}
\begin{equation}
          \bm{X}_{\mathrm{d}}^{n} = \mathrm{Downsample}({\bm{X}_{\mathrm{in}}^{n}}),
\end{equation}
where $\mathrm{Downsample}(\cdot)$ is a weighted pooling operation among the neighboring tokens, and $\mathrm{Linear}(\cdot)$ is a linear layer. $\mathrm{TM}(\cdot)$ is a temporal merge block that uses a pooling operation to shrink the number of interaction feature tokens. 
 
Similarly, to explore the temporal relationships, we adopt the same operation on the query set videos but do not utilize category names. For the whole PSTI module, we repeat the above process to recurrently inject velocity-tailored text information and progressively interact with video-text information at different velocity levels. 

\subsection{Multi-Velocity Feature Alignment}
\label{multi-velocity}
Given the substantial variations in motion speed among action instances, it becomes possible to simultaneously perform multi-velocity feature matching for each specific action. As mentioned in the above section, we acquire multiple velocity representations, with each carrying distinct velocity information. Consequently, we propose a Multi-Velocity Feature Alignment (MVFA) module to capture pair-wise semantic relevance at different velocity levels, thereby facilitating coherent matching between videos.

Specifically, given different velocity action feature outputs from the PSTI module, the proposed MVFA module calculates the query-support distance at each velocity level and then obtains the overall query-support distance by the weighted sum of the distances. Specifically, given the support visual feature outputs $\bar{\bm{X}^{n}_{\mathrm{s}}}=[\bar{\bm{x}}^{{\prime}1}_{\mathrm{s}},...,\bar{\bm{x}}^{{\prime}{2^{4-n}}}_{\mathrm{s}} ]$ and the query visual features $\bar{X^{n}_{\mathrm{q}}}=[\bar{\bm{x}}^{{\prime}1}_{\mathrm{q}},...,\bar{\bm{x}}^{{\prime}{2^{4-n}}}_{\mathrm{q}} ]$ at velocity scale $n$, we adopt the temporal alignment metric on these features:
\begin{equation}
    \mathrm{D}(\bar{\bm{X}^{n}_{\mathrm{s}}}, \bar{\bm{X}^{n}_{\mathrm{q}}}) = \mathcal{M} ([\bar{\bm{x}}^{{\prime}1}_{\mathrm{s}},...,\bar{\bm{x}}^{{\prime}{2^{4-n}}}_{\mathrm{s}} ], [\bar{\bm{x}}^{{\prime}1}_{\mathrm{q}},...,\bar{\bm{x}}^{{\prime}{2^{4-n}}}_{\mathrm{q}}]),
\end{equation}
where $\mathrm{D}(\bar{\bm{X}^{n}_{\mathrm{s}}}, \bar{\bm{X}^{n}_{\mathrm{q}}})$ is the distance between the query visual feature $\bar{\bm{X}^{n}_{\mathrm{q}}}$ and the support visual feature $\bar{\bm{X}^{n}_{\mathrm{s}}}$. $\mathcal{M}$ denotes the OTAM temporal metric~\cite{cao2020few} by default in our MVP-Shot. The proposed MVP-Shot is designed as a plug-and-play framework. We will incorporate MVP-Shot into other existing metrics in the following experimental section, and empirically demonstrate its pluggability. Furthermore, we express the total support-query distance between support and query videos as the weighted sum of the distances calculated across multiple velocity scales:
\begin{equation}\label{eq: distance}
    \mathrm{D}_{\mathrm{s},\mathrm{q}} = \sum_{n=1}^{N}{{\alpha}_{n} \mathrm{D}(\bar{\bm{X}^{n}_{\mathrm{s}}}, \bar{\bm{X}^{n}_{\mathrm{q}}})},
\end{equation}
where ${\alpha}_{n}$ is the coefficient.  The maximum velocity scale $N$ is set to $3$ by default in this paper for considering the balance between effectiveness and efficiency.

To tackle the issue that different action instances may be performed at different motion dynamics, we propose PSTI and MVFA to progressively capture and align multi-velocity action features. By this means, our proposed MVP-shot combines two modules to complement each other to make more accurate query predictions under the few-shot setting.

Finally, we can use the output total query-support distance $\mathrm{D}_{\mathrm{s},\mathrm{q}}$ as logits for further classification. Following previous works~\cite{wang2022hybrid,boosting,wang2023molo}, we can optimize the model parameters using a cross-entropy loss $\mathcal{L}_{CE}$. For few-shot inference, we can obtain final prediction results based on the nearest neighbor rule~\cite{snell2017prototypical} by utilizing the distance values of support-query video pairs in Eq.\eqref{eq: distance}.

\begin{table*}[ht]
\caption{Comparison results with recent  state-of-the-art few-shot action recognition methods (\S\ref{sota}) on  Kinetics~\cite{carreira2017quo}, UCF101~\cite{soomro2012ucf101} and HMDB51~\cite{kuehne2011hmdb}. We conduct experiments under the $5$-way $K$-shot settings. We highlight the best results. "INet-RN50" denotes ResNet-50 visual encoder.}
\label{table1}
\centering
\resizebox{0.96\textwidth}{!}{
\renewcommand{\arraystretch}{1.4}
\setlength\tabcolsep{3pt}
\begin{tabular}{rl|c||ccc|ccc|ccc}
\thickhline \rowcolor{mygray}
      & &  & \multicolumn{3}{c|}{Kinetics} & \multicolumn{3}{c|}{UCF101} & \multicolumn{3}{c}{HMDB51} \\ 
  \rowcolor{mygray}   \multicolumn{2}{c|}{\multirow{-2}{*}{Method}}    &       \multirow{-2}*{Pre-training}      &$1$-shot (\%)    & $3$-shot (\%)   & $5$-shot (\%)   & $1$-shot (\%)   & $3$-shot (\%)   & $5$-shot (\%)  & $1$-shot (\%)   & $3$-shot (\%)  & $5$-shot (\%)  \\ \hline
 \hline
MatchingNet~\cite{vinyals2016matching}\!\!&\pub{NeurIPS16}  &  INet-RN50 & $53.3$     & $69.2$    & $74.6$    & -       & -       & -      & -       & -       & -      \\
CMN~\cite{zhu2018compound}\!\!&\pub{ECCV18}        &  INet-RN50   & $57.3$     & $75.6$    & $76.0$      & -       & -       & -      & -       & -       & -      \\
ARN~\cite{zhang2020few}\!\!&\pub{ECCV20}       & INet-RN50  & $63.7$     & -       & $82.4$   & $66.3$    & -       & $83.1$   & $45.5$    & -       & $60.6$   \\
OTAM~\cite{cao2020few}\!\!&\pub{CVPR20}       &  INet-RN50 & $73.0$       & $78.7$    & $85.8$    & $79.9$    & $87.0$      & -      & $54.5$    & $65.7$    & -      \\
ITANet~\cite{zhang2021learning}\!\!&\pub{IJCAI21}       &  INet-RN50 & $73.6$       & -    & $84.3$    & -    & -    & -      & -    & -    & -      \\

TRX~\cite{perrett2021temporal}\!\!&\pub{CVPR21}     &  INet-RN50   & $63.6$     & $81.8$    & $85.9$    & $78.2$    & $92.4$    & $96.1$   & $53.1$    & $66.8$    & $75.6$   \\

TA$^{2}$N~\cite{ta2n}\!\!&\pub{AAAI22}       &  INet-RN50   & $72.8$     & -       & $85.8$    & $81.9$    & -       & $95.9$   & $59.7$      & -       & $73.9$   \\

MTFAN~\cite{wu2022motion}\!\!&\pub{CVPR22}      &  INet-RN50   & $74.6$     & -       & $87.4$    & $84.8$    & -       & $95.1$   & $59.0$      & -       & $74.6$   \\

HyRSM~\cite{wang2022hybrid}\!\!&\pub{CVPR22}     &  INet-RN50  & $73.7$     & $83.5$    & $86.1$    & $83.9$    & $93.0$      & $94.7$   & $60.3$    & $71.7$    & $76.0$     \\

STRM~\cite{thatipelli2022spatio}\!\!&\pub{CVPR22}       &  INet-RN50   & $62.9$     & $81.1$    & $86.7$    & $80.5$    & $92.7$    & $96.9$   & $52.3$    & $67.4$    & $77.3$   \\
CPM~\cite{huang2022compound}\!\!&\pub{ECCV22}       & INet-RN50  & $73.3$     & -       & -       & $71.4$    & -       & -      & $60.1$    & -       & -      \\
HCL~\cite{zheng2022few}\!\!&\pub{ECCV22}         &  INet-RN50    & $73.7$     & $82.4$    & $85.8$    & $82.6$    &$91.0$      & $94.5$   & $59.1$    & $71.2$    & $76.3$   \\

MoLo~\cite{wang2023molo}\!\!&\pub{CVPR23}       & INet-RN50    & $74.0$       & $83.7$    & $85.6$    & $86.0$      & $93.5$   & $95.5$   & $60.8$    & $72.0$      & $77.4$   \\ 
GgHM~\cite{boosting}\!\!&\pub{ICCV23}       & INet-RN50    & $74.9$       & $-$    & $87.4$    & $85.2$      & $-$   & $96.3$   & $61.2$    & $-$      & $76.9$   \\ 
CLIP-Freeze~\cite{radford2021learning}\!\!&\pub{ICML21}      &  CLIP-RN50    & $68.2$       & $82.1$    & $85.3$    & $84.6$      & $92.9$    & $94.5$   & $51.4$    & $65.2$      & $71.0$   \\ 
CLIP-FSAR~\cite{wang2023clip}\!\!&\pub{IJCV23}      &  CLIP-RN50    & $87.6$       & $90.7$    & $91.9$    & $91.3$      & $95.1$    & $97.0$   & $69.2$    & $77.6$      & $80.3$   \\ 
\hline \hline
 \multicolumn{2}{c|}{MVP-shot  \textbf{(Ours)}}         & CLIP-RN50       & $\mathbf{90.0}$    & $\mathbf{92.0}$   & $\mathbf{93.2}$   & $\mathbf{92.2}$    & $\mathbf{96.9}$  & $\mathbf{97.6}$   & $\mathbf{72.5}$    & $\mathbf{79.7 }$   & $\mathbf{82.5}$\\ \thickhline
\end{tabular}}
\end{table*}
\begin{table}[!t]
\caption{Comparison results with recent  state-of-the-art few-shot action recognition methods (\S\ref{sota}) on  SSv2-small~\cite{goyal2017something} dataset. We conduct experiments under the $5$-way $K$-shot settings. We highlight the best results. "INet-RN50" denotes ResNet-50 visual encoder.}
\label{table2}
\centering
\resizebox{0.48\textwidth}{!}{
\renewcommand{\arraystretch}{1.2}
\setlength\tabcolsep{0.6pt}
\begin{tabular}{rl|c||ccc}
\thickhline
\rowcolor{mygray} &    &  & \multicolumn{3}{c}{\textbf{SSv2-small}} \\
   \rowcolor{mygray}  \multicolumn{2}{c|}{\multirow{-2}{*}{Method}}       &\multirow{-2}*{\textbf{Pre-training}}            & $1$-shot (\%)   & $3$-shot (\%)   & $5$-shot (\%)   \\ \hline \hline
MatchingNet~\cite{vinyals2016matching}\!&\pub{NeurIPS16} &  INet-RN50 & $31.3$    & $39.8$    & $45.5$    \\ 
MAML~\cite{finn2017model}\!&\pub{ICML17}     & INet-RN50  & $30.9$    & $38.6$    & $41.9$    \\ 
CMN~\cite{zhu2018compound}\!&\pub{ECCV18}        & INet-RN50  & $36.2$    & $42.5$    & $48.8$    \\ 
OTAM~\cite{cao2020few}\!&\pub{CVPR20}     &  INet-RN50  & $36.4$    & $45.9 $   & $48.0$      \\ 
ITANet~\cite{zhang2021learning}\!&\pub{IJCAI21}      & INet-RN50  & $39.8$    & $49.4 $      & $53.7$    \\ 
TRX~\cite{perrett2021temporal}\!&\pub{CVPR21}       &  INet-RN50   & $36.0$     & $51.9$    & $56.7$ \\ 
HyRSM~\cite{wang2022hybrid}\!&\pub{CVPR22}      &  INet-RN50  & $40.6$    & -       & $56.1$    \\ 
STRM~\cite{thatipelli2022spatio}\!&\pub{CVPR22}     &  INet-RN50  & $37.1$    & $49.2 $   & $55.3$    \\ 
HCL~\cite{zheng2022few}\!&\pub{ECCV22}       &  INet-RN50  & $38.9 $   & $49.1$    & $55.4$    \\ 
MoLo~\cite{wang2023molo}\!&\pub{CVPR23}      &  INet-RN50  & $42.7$    & $50.9 $   & $56.4$    \\ 
CLIP-Freeze~\cite{radford2021learning}\!&\pub{ICML21}      &  CLIP-RN50   &  $26.8$   & $33.7$   & $36.3$   \\
CLIP-FSAR~\cite{wang2023clip}\!&\pub{IJCV23}      &  CLIP-RN50   &  $\mathbf{52.0}$   & $54.0$   & $55.8$   \\   \hline \hline

 \multicolumn{2}{c|}{MVP-shot  \textbf{(Ours)}}              & CLIP-RN50   & $51.2 $   &  $\mathbf{55.2}$   & $\mathbf{57.0}$ \\ \thickhline
\end{tabular}}
\vspace{-0.08in}
\end{table}
\section{Experiments}
\label{experiments}
In this section, we first introduce four few-shot action recognition datasets (~\ie, UCF101~\cite{soomro2012ucf101}, Kinetics~\cite{carreira2017quo}, HMDB51~\cite{kuehne2011hmdb}, and SSv2-small~\cite{goyal2017something}) and the evaluation protocol (\S\ref{setup}). Then we describe the implementation details of our MVP-shot (\S\ref{details}). Following that, we benchmark our methods against previous state-of-the-art few-shot action recognition approaches on four widely used datasets (\S\ref{sota}). Finally, to demonstrate the effectiveness of the proposed MVP-shot, we conduct several ablation experiments(\S\ref{ablatio}).
\subsection{Experimental Setup}
\label{setup}
\sssection{Dataset.} We evaluate the effectiveness of our proposed MVP-shot on four benchmarks,~\ie, Kinetics~\cite{carreira2017quo}, SSv2-small~\cite{goyal2017something}, UCF101~\cite{soomro2012ucf101}, and HMDB51~\cite{kuehne2011hmdb}. We evaluate all models on different datasets using $5$-way $1$/$3$/$5$-shot few-shot recognition tasks. For SSv2-small and Kinetics, we utilize the split setting as in  CMN~\cite{zhu2018compound}, which both randomly select $64$ classes from the original dataset as the \texttt{train} set, $12$ classes as the \texttt{val} set and $24$ classes as the \texttt{test} set. The  UCF101\cite{soomro2012ucf101} dataset consists of 101 action classes, and we utilize the splits as in \cite{zhang2020few}, where there are 70, 10, and 21 classes for \texttt{train/val/test}, respectively. For HMDB51\cite{kuehne2011hmdb}, we utilize the splits from \cite{zhang2020few}, and the dataset has 31 action classes, 10 action classes, and 10 action classes for \texttt{train/val/test}, respectively. 

\sssection{Evaluation.} Following the official evaluation protocols~\cite{boosting,cao2020few}, we use $5$-way $1$-shot, $3$-shot  and  $5$-shot accuracy for evaluation. We randomly select $10000$ episodic tasks from \texttt{test} and report average results.

\subsection{Implementation Details}
\label{details}
\sssection{Network Architecture.} We utilize pre-trained CLIP ResNet-50 as our basic visual encoder to ensure a fair comparison with previous methods~\cite{cao2020few,perrett2021temporal}, which choose Resnet-50~\cite{he2016deep} pre-trained on ImageNet~\cite{deng2009imagenet} as the visual encoder. In order to make full use of the pre-trained prior knowledge in CLIP and alleviate the optimization burden, we freeze the text encoder and prevent it from being updated during training.  

\sssection{Network Training.} Following previous methods~\cite{wang2016temporal}, we sparsely and uniformly sample $T$ = $8$ frames of the raw video sequences to encode the frame-level video representation. Each video frame is scaled to a height of $256$. Then we randomly crop these resized video frames into the size of 224 × 224. In the training phase, following previous methods~\cite{wang2022hybrid,wang2023clip}, we employ basic data augmentation, such as color jitter, random horizontal flipping, and cropping. Moreover, we use the Adam optimizer with the multi-step scheduler to train our model. By default, we set the total number of training steps to $10$. In the inference phase, we use the average of the 10000 episodes as the result. For many shot classifications, we follow the prototype paradigm~\cite{snell2017prototypical} and acquire the mean features of support videos within each class to serve as prototypes, so as to calculate the distance between prototypes and query videos for classification. To guarantee reproducibility, full code will be released.

\subsection{Comparison with SOTA methods}
\label{sota}
\textbf{Results on Kintices~\cite{carreira2017quo}, HMDB51~\cite{kuehne2011hmdb} and UCF101~\cite{soomro2012ucf101}.} AS shown in Table ~\ref{table1}, we compare our MVP-shot with current state-of-the-art few-shot action recognition on three standard datasets. Table ~\ref{table1} demonstrates our method outperforms other state-of-the-art methods on  Kinetics, HMDB51, and UCF101 datasets under different shot settings. Specifically, our model has achieved significant improvements in accuracy under any task setting compared to previous excellent methods (such as MoLo~\cite{wang2023molo}, HyRSM~\cite{wang2022hybrid}, and GgHM~\cite{boosting}). Compared to CLIP-FSAR~\cite{wang2023clip}, which makes use of semantic information to perform cross-modal interaction, our MVP-shot achieves better results in multiple datasets in that our method performs multi-velocity feature matching. Notably, our method achieves an increase of $3.3$\%, $2.4$\% performance improvements on HMDB51 and Kinetics datasets under the 1-shot task, and achieves an increase of $2.2$\%, $1.3$\% performance improvements on HMDB51 and Kinetics datasets under the 5-shot task compared to CLIP-FSAR~\cite{wang2023clip}, respectively.
 
\textbf{Results on SSv2-small~\cite{goyal2017something}.} We also report our MVP-shot model results on SSV2-small datasets on Table~\ref{table2}. Our MVP-shot achieves better performance in the $3$-shot and $5$-shot tasks than all other state-of-the-art methods on SSV2-small datasets. Compared to CLIP-FSAR~\cite{wang2023clip}, our method also has a significant performance improvement. Concretely,  our MVP-shot brings $1.0$\% performance improvements in the 3-shot task, and $1.2$\% performance improvements under the 5-shot setting, respectively.  
\begin{table}[!t]
\caption{Ablation study of each component in our MVP-shot (\S\ref{ablatio}). We reported the 1-shot and 5-shot results on Kinetics~\cite{carreira2017quo} dataset. The best results are highlighted.}
\label{table3}
\centering
\resizebox{0.38\textwidth}{!}{
\setlength\tabcolsep{8pt}
\renewcommand\arraystretch{1.2}
\begin{tabular}{cc||cc}
\thickhline
\rowcolor{mygray} 
&  & \multicolumn{2}{c}{Kinetics} \\
\rowcolor{mygray} \multirow{-2}*{PSTI} &\multirow{-2}*{MVFA} & $1$-shot (\%) & $5$-shot (\%) \\ \hline\hline
  \xmark  &  \xmark        & $79.8$  & $89.3$   \\ 
   \cmark   &   \xmark      & 88.3    & $92.3$     \\ 
     \xmark   & \cmark        & $81.9$  & $91.8$   \\ 
 \cmark      & \cmark           & $\mathbf{90.0}$   & $\mathbf{93.2}$   \\ \hline
\end{tabular}}
\end{table}

\begin{table}[!t]
\caption{Ablation study of velocity scales in multi-velocity feature alignment in our MVP-shot (\S\ref{ablatio}). We reported the 1-shot and 5-shot results on Kinetics~\cite{carreira2017quo} dataset. The best results are highlighted.}
\label{table4}
\centering
\resizebox{0.38\textwidth}{!}{
\setlength\tabcolsep{4pt}
\renewcommand\arraystretch{1.4}
\begin{tabular}{ccc||cc}
\thickhline
\rowcolor{mygray} 
\multicolumn{3}{c||}{Velocity}  &  \multicolumn{2}{c}{Kinetics}\\ 
 \rowcolor{mygray}   $n=1$        &      $n=2$     & $n=3$&   $1$-shot (\%)    &  $5$-shot (\%)  \\
\hline \hline
 \xmark    &\xmark &    \cmark       &  $81.6$ &  $87.0$  \\
      \xmark     & \cmark &   \xmark       &   $86.0$ & $90.5$  \\  
      \cmark      & \xmark &  \xmark      &  $88.3$   &   $92.3$ \\
            \xmark   & \cmark  & \cmark       & $81.9$  & $91.8$ \\
            \cmark     &\xmark &  \cmark        & $88.8$    & $92.5$     \\ 
   \cmark      & \cmark &   \xmark       & $89.5$  & $93.0$   \\ 

\cmark     &  \cmark   &  \cmark       & $\mathbf{90.0}$   & $\mathbf{93.2}$   \\ \hline
\end{tabular}}
\end{table}

Our MVP-shot achieves the best performance on all datasets, which reveals that our MVP-shot has strong generalization across diverse datasets. In addition, we can observe the following from the experimental results: (1) Compared with CLIP-FSAR~\cite{wang2023clip}, which makes use of semantic information to perform frame-level feature matching, our MVP-shot achieves better performance in multiple task settings and datasets. Specifically, MVP-shot brings $3.3$\% and $2.4$\% performance improvements on HMDB51 and Kinetics under the 1-shot setting, respectively.  It indicates our MVP-shot further boosts performance by exploring multi-velocity feature matching. (2) We observe that MVP-shot achieves better performance than current state-of-the-art methods (such as HyRSM~\cite{wang2022hybrid}, TRX~\cite{perrett2021temporal}, and MoLo~\cite{wang2023molo}), especially under the $1$-shot and $3$-shot settings. This indicates that incorporating textual semantic information would be more effective for enhancing semantic-related velocity features when visual information is insufficient.

\subsection{Ablation Study}
\label{ablatio}
We conduct ablation experiments to substantiate the efficacy of our proposed MVP-shot and dissect the contribution of each component. Unless explicitly stated otherwise, we adopt the CLIP Resnet-50 model as our visual encoder by default.

\noindent \textbf{Analysis of Each Module in MVP-shot.} 
To validate the impacts of each module in our MVP-shot, we conduct experiments on the Kinetics~\cite{carreira2017quo} under $5$-way $1$-shot and $5$-shot settings. Table~\ref{table3} summarizes the effects of each module in MVP-shot.  By performing multi-velocity feature alignment (MVFA) (\cf~\S\ref{progressive}), we obtain $2.1$\% and $2.5$\% performance gains under $1$-shot and $5$-shot settings, respectively. The consistent promotion indicates that MVFA can capture pair-wise semantic relevance at different velocity scales to facilitate accurate matching between videos. In addition, the proposed Progressive Semantic-Tailored Interaction (PSTI) module (\cf~\S\ref{multi-velocity}) improves by $8.5$\% and $3.0$\% on $1$-shot and $5$-shot settings, respectively, which indicates recurrently injecting velocity-tailored text information significantly boosts performance. We observe a notable performance improvement, particularly in the 1-shot scenario, indicating that the supplementation of textual semantic information proves more effective when visual information is severely limited. Moreover, stacking MVFA and PSTI modules can compensate for each other to further improve performance.
\begin{table}[!t]
\caption{Comparison of different feature interaction methods (\S\ref{ablatio}). We reported the $1$-shot and $5$-shot results on Kinetics~\cite{carreira2017quo} dataset.}
\label{table5}
\centering
\resizebox{0.38\textwidth}{!}{
\setlength\tabcolsep{3pt}
\renewcommand\arraystretch{1.2}
\begin{tabular}{l||cc}
\thickhline
\rowcolor{mygray}
 & \multicolumn{2}{c}{Kinetics} \\
\rowcolor{mygray}
\multirow{-2}*{Method}&$1$-shot (\%) & $5$-shot (\%) \\ \hline\hline
   Single-modal Transformer       & $81.5$  & $92.0$   \\
   Multi-modal Transformer     & $88.9$    & $92.7$     \\ 
     \textbf{STI (ours)} & $\mathbf{90.0}$   & $\mathbf{93.2}$   \\ \hline
\end{tabular}}
\end{table}

\noindent \textbf{Analysis of Multi-Velocity Feature Alignment.} 
To demonstrate that multi-velocity feature alignment coordinates with each other to facilitate accurate matching between videos, we conduct the experiments on Kinetics~\cite{carreira2017quo} under the $5$-way $1$-shot and $5$-shot settings by various combinations of multi-velocity feature alignment. We report the obtained results in Table~\ref{table4}. When the action feature is provided at a single velocity, the corresponding performance is reduced on the $1$-shot and $5$-shot settings. The performance of multi-velocity feature alignment improves as the velocity scales increase. Further, our MVP-shot, which combines all velocity feature alignments, achieves the best performance. Specifically, MVP-shot outperforms all single-velocity feature alignments by $1.7$\%, $4.0$\%, and $8.4$\% on the $1$-shot setting, respectively. These results demonstrate that our proposed multi-velocity feature alignment can surpass corresponding methods with single-velocity feature alignment, and capture pair-wise semantic relevance at different velocity scales to facilitate coherent matching between videos.

\begin{figure}[!t]
    \centering
    \includegraphics[width=0.8\linewidth]{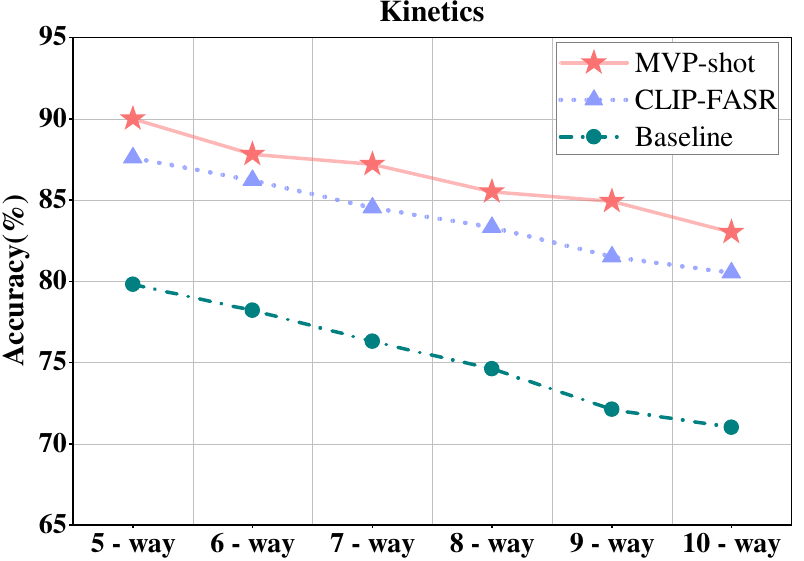}
    \caption{$N$-way $1$-shot results of our MVP-shot on Kinetics~\cite{carreira2017quo} dataset, our baseline methods and
CLIP-FSAR with $N$ varying from $5$ to $10$.}
    \label{fig4}
    \vspace{-0.05in}
\end{figure}

\noindent \textbf{Comparison with Different Feature Interaction Methods}. To evaluate the efficacy of different feature interaction methods and demonstrate the superiority of our proposed Semantic-Tailored Interaction (STI) module, we perform comparative analysis on Kinetics~\cite{carreira2017quo} under the 5-way 1-shot and 5-way 5-shot settings. Specifically, we replace our STI module with several other feature interaction methods. The transformer\cite{transformer} consists of the multi-head self-attention multi-head self-attention mechanisms and a feed-forward network. The single-modal transformer indicates the support features and query features performing self-attention on the temporal dimension without text features. Compared with the first baseline, the second multi-modal transformer ( CLIP-FSAR~\cite{wang2023clip} ) stacks the support features to corresponding textual features along the temporal dimension before self-attention. All the layers of the transformer are set to be one. All other experimental settings remain consistent with those of the proposed MVP-shot to ensure a fair comparison. We report the experimental results in Table ~\ref{table5}. In comparison, the proposed STI module outperforms the unimodal transformer by $8.5$\% ($90.0$\% vs. $81.5$\%) and $1.2$\% ($93.2$\% vs. $92.0$\%) under the $1$-shot and $5$-shot settings on Kinetics dataset, respectively. The observed performance gains suggest complementarity between textual and visual modalities, wherein semantic information augments the discriminative capacity of visual features. Compared with the multimodal transformer, the proposed STI module increases  $1.7$\% ($88.3$\% vs. $90.0$\%) and $0.9$\% ($92.3$\% vs. $93.2$\%) under the $1$-shot and $5$-shot settings on Kinetics dataset, respectively. Experimental results demonstrate that STI module enhances the utilization of textual information more efficiently to capture the semantic-related velocity features. Thus, our proposed module can achieve more robust and more accurate class prototype estimation.
\begin{table}[!t]
\caption{Generalization performance of MVP-shot with different temporal alignment metrics (\S\ref{ablatio}) on  Kinetics~\cite{carreira2017quo} under the $5$-way $1$-shot and $5$-way $5$-shot settings.}
\label{table6}
\centering
\resizebox{0.38\textwidth}{!}{
\setlength\tabcolsep{4pt}
\renewcommand\arraystretch{1.2}
\begin{tabular}{l||cc}
\thickhline
\rowcolor{mygray}
  & \multicolumn{2}{c}{Kinetics} \\
 \rowcolor{mygray}\multirow{-2}*{Method}& $1$-shot (\%) & $5$-shot (\%) \\ \hline \hline
   Bi-MHM~\cite{wang2022hybrid}       & $81.5$ &  $91.6$  \\
   CLIP-FSAR (Bi-MHM)       & $87.7$ &  $92.1$  \\ 
   \textbf{Ours (Bi-MHM)}      & $\mathbf{89.4}$  & $\mathbf{92.9}$   \\  \hline
   OTAM~\cite{cao2020few}    & $81.9$  & $91.8$   \\
   CLIP-FSAR (OTAM)       & $87.6$ &  $91.9$  \\ 
       \textbf{Ours (OTAM)}  & $\mathbf{90.0}$   & $\mathbf{93.2}$   \\
       \hline
\end{tabular}}
\end{table}
\begin{table*}[ht]
\caption{Influence of different backbones. Quantitative comparison results (\S\ref{ablatio}) on Kinetics~\cite{carreira2017quo}, UCF101~\cite{soomro2012ucf101}, HMDB51~\cite{kuehne2011hmdb}, and SSv2-small~\cite{goyal2017something}. We highlight the best results.}
\label{table7}
\centering
\resizebox{\textwidth}{!}{
\renewcommand{\arraystretch}{1.3}
\setlength\tabcolsep{2pt}
\begin{tabular}{l|c||cc|cc|cc|cc}
\thickhline \rowcolor{mygray}
      &  & \multicolumn{2}{c|}{Kinetics} & \multicolumn{2}{c|}{UCF101} & \multicolumn{2}{c|}{HMDB51} & \multicolumn{2}{c}{SSv2-small} \\ 
  \rowcolor{mygray}  \multirow{-2}{*}{Method}   &       \multirow{-2}*{Pre-training}      &$1$-shot (\%)       & $5$-shot (\%)   & $1$-shot (\%)   & $5$-shot (\%)   & $1$-shot (\%)  & $5$-shot (\%)   & $1$-shot (\%)  & $5$-shot (\%)  \\ \hline
 \hline
CLIP-FSAR  &   \multirow{2}{*}{CLIP-RN50} & $87.6$     & $91.9$    & $91.3$    & $97.0$       & $69.2$       & $80.3$      & $52.0$       & $55.8$           \\
\textbf{MVP-shot (Ours)}       &    & $\mathbf{90.0}$      & $\mathbf{93.2}$   & $\mathbf{92.2}$      & $\mathbf{97.6}$   & $\mathbf{72.5}$    &    $\mathbf{82.5}$     & $\mathbf{51.2}$     & $\mathbf{57.0}$ \\ \hline
CLIP-FSAR      & \multirow{2}{*}{CLIP-ViT-B}      & $89.7$ & $95.0$  &  $96.6$ & $99.0$&$75.8$ & $87.7$  & $54.5$   &  $61.8$        \\
\textbf{MVP-shot (Ours)}    &       &  $\mathbf{91.0}$   & $\mathbf{95.1}$ & $\mathbf{96.8}$& $\mathbf{99.0}$ & $\mathbf{77.0}$ & $\mathbf{88.1}$ &  $\mathbf{55.4}$   & $\mathbf{62.0} $    \\\hline

\end{tabular}}
\end{table*}

\noindent \textbf{Influence$_{\!}$ of$_{\!}$ Different$_{\!}$ Alignment$_{\!}$ Strategies.$_{\!}$} We utilize different temporal alignment metrics to conduct experiments on Kinetics~\cite{carreira2017quo} under the $5$-way $1$-shot and $5$-shot settings to demonstrate our model generalizability.$_{\!}$ Different temporal alignment metrics are adopted, including OTAM~\cite{cao2020few} and Bi-MHM~\cite{wang2022hybrid}. As shown in Table~\ref{table6}, our MVP-shot is adaptable to any temporal alignment metric and achieves great performance.  Moreover, irrespective of employed temporal alignment metric, compared to the naive baselines, our framework still achieves significant performance gains, indicating the superiority of our MVP-shot.

\noindent \textbf{Influence of Different Backbones.} The previous comparison results we reported are all based on CLIP-RN50 visual encoder. Here, we conduct detailed experiments to evaluate the influence of different backbones. We adopt the CLIP-ViT-B backbone, and the other settings are the same as those of our work. We list comparison results with CLIP-FSAR based on the CLIP-ViT-B backbone in Table~\ref{table7}. We conduct experiments on four widely used benchmarks (\ie, HMDB51~\cite{kuehne2011hmdb}, UCF101~\cite{soomro2012ucf101}, kinetics100~\cite{carreira2017quo}, and SSv2-small~\cite{goyal2017something}). From Table~\ref{table7}, we can see based on CLIP-ViT-B, our MVP-shot consistently outperforms CLIP-FSAR~\cite{wang2023clip}, demonstrating the superiority of our method. In addition, our MVP-shot model utilizing CLIP-ViT-B generally outperforms its counterpart based on CLIP-RN50, suggesting that a more robust pre-trained visual model leads to enhanced few-shot generalization.

\noindent \textbf{Testing of $N$-Way Few-Shot Action Recognition.} 
We carry out all of the comparative evaluation experiments under the $5$-way setting in the previous experiments. To further investigate the effect of varying categories $N$  on the few-shot performance, we compare $N$-way $1$-shot results on Kinetics (as shown in Fig.~\ref{fig4}). The results indicate that as the value of $N$ increases, the classification accuracy gradually becomes more challenging. For instance, the $10$-way performance of MVP-shot exhibits a decrease of $7.0$\% compared to the 5-way scenario. Nevertheless, the performance of our MVP-shot is still consistently better than the comparison methods, such as CLIP-FSAR~\cite{wang2023clip}. This demonstrated that our proposed MVP-shot can boost performance by jointly exploring multi-velocity feature representation and multi-velocity feature alignment. 

\begin{figure}[!t]
\vspace{-0.1in}
    \centering
    \includegraphics[width=0.8\linewidth]{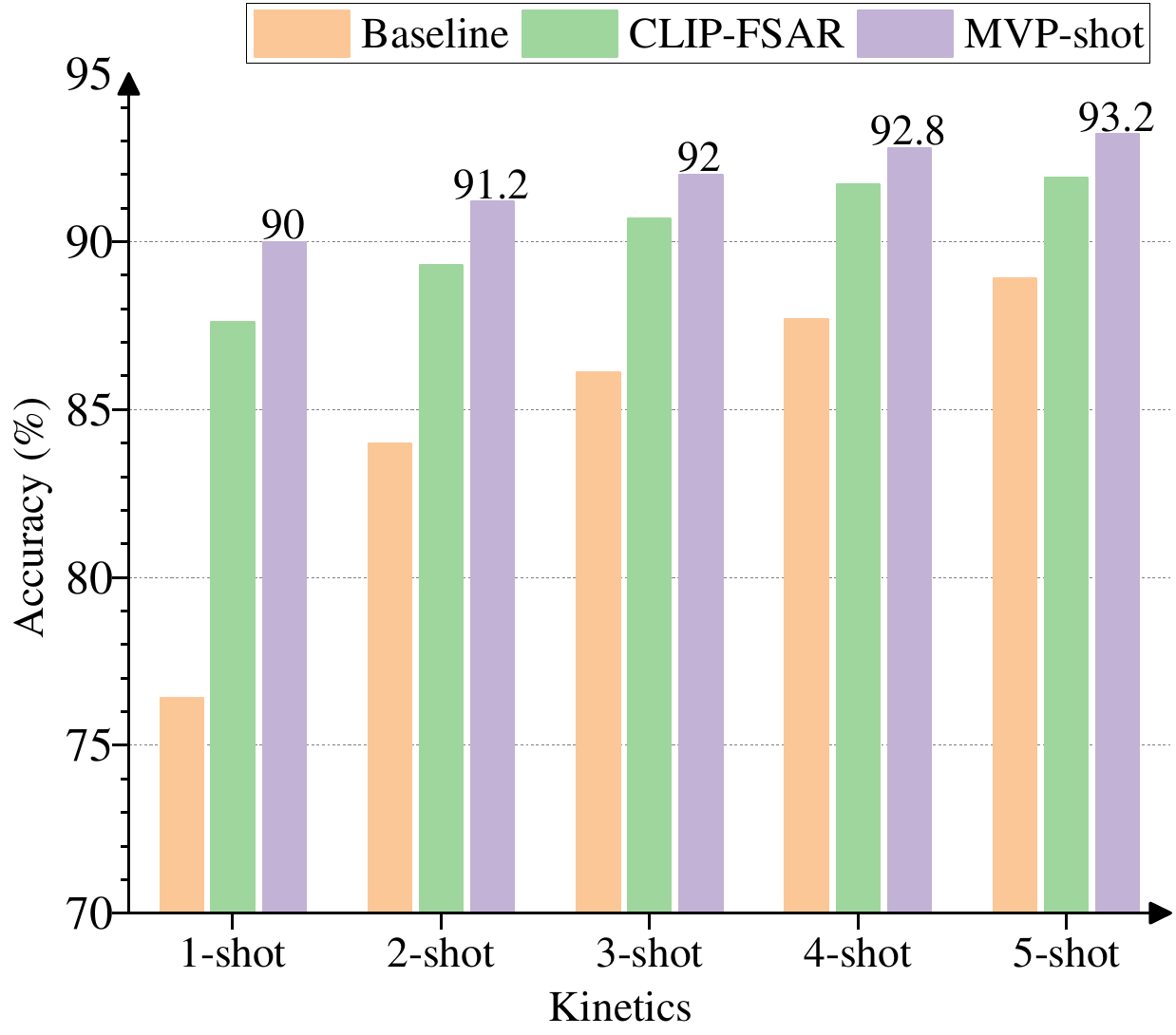}
    \caption{ Comparison results (\S\ref{ablatio}) with different numbers of support
samples on kinetics~\cite{carreira2017quo} dataset under $5$-way $K$-shot settings.}
    \label{sup:kshot}
    \vspace{-0.2in}
\end{figure}

\noindent \textbf{$5$-way $K$-shot Few-shot Classification.}
To more comprehensively assess our model's performance across diverse few-shot scenarios, we conduct additional experiments by progressively increasing the number of support samples from $1$-shot to $5$-shot on Kinetics~\cite{carreira2017quo}. As $K$ decreases, the classification difficulty becomes higher, and the performance decreases (as shown in Fig.~\ref{sup:kshot}). Though the performance of MVP-shot and CLIP-FSAR~\cite{wang2023clip} is limited due to too few support samples, MVP-shot performs consistently better than baseline and CLIP-FSAR in all settings. This demonstrates that our proposed MVP-shot can boost performance by exploring multi-velocity feature matching. 

\begin{figure*}[!t]
    \centering
    \includegraphics[width=0.98\textwidth]{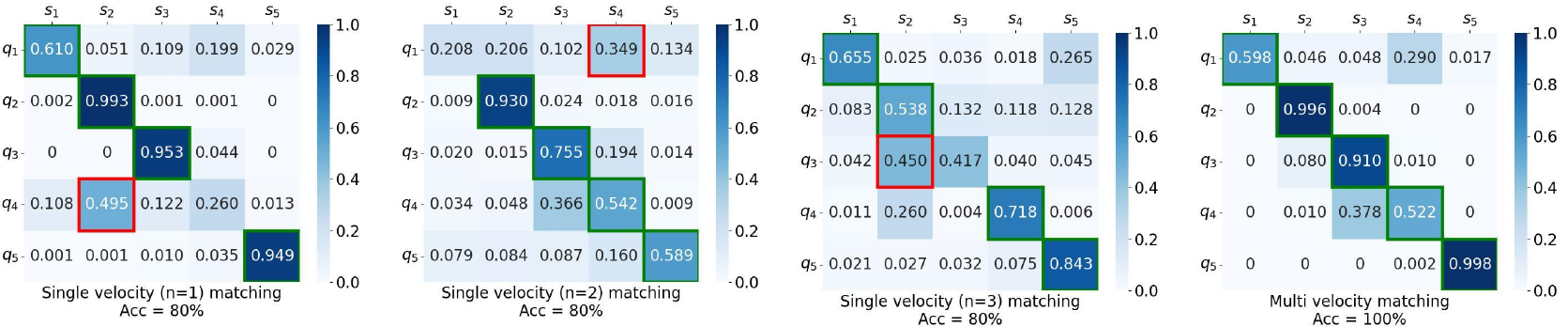}
    \caption{Similarity visualization between support prototypes ($s_{n}$) and query samples ($q_{n}$)  with different methods in an episode task sampled from Kinetics. The action classes, from left to right, are \textit{Shot Put}, \textit{Catching or Throwing Baseball}, \textit{Playing Ice Hockey}, \textit{Drop Kicking} and \textit{Dribbling Basketball}, respectively. The correct prediction is indicated by the {\color{green} green} box, while the {\color{red} red} box denotes an incorrect prediction.}
    \label{fig5}
    \vspace{-0.05in}
\end{figure*}
\subsection{Visualization Analysis}
\noindent \textbf{Similarity Visualization.} To qualitatively demonstrate the proposed multi-velocity feature alignment in our MVP-shot, we visualize the predicted similarities between query videos and class prototypes for an episode under both single-velocity and multi-velocity settings in Fig.~\ref{fig5}. Compared to single-velocity alignment, multi-velocity alignment captures pair-wise semantic relevance at different velocity scales, compensating for the shortcomings of single-velocity alignment. Thus, our proposed multi-velocity alignment makes more accurate decisions. Notably, for the fourth query sample from Kinetics, the erroneous decision made by the single-velocity (n=$1$) matching can be rectified by the fusion of multi-velocity matching results. Experimental results demonstrate that multi-velocity matching Complements each other, and combines alignment across all velocity features, achieving the best performance.

\begin{figure*}[!t]
    \centering
   \includegraphics[width=0.8\textwidth]{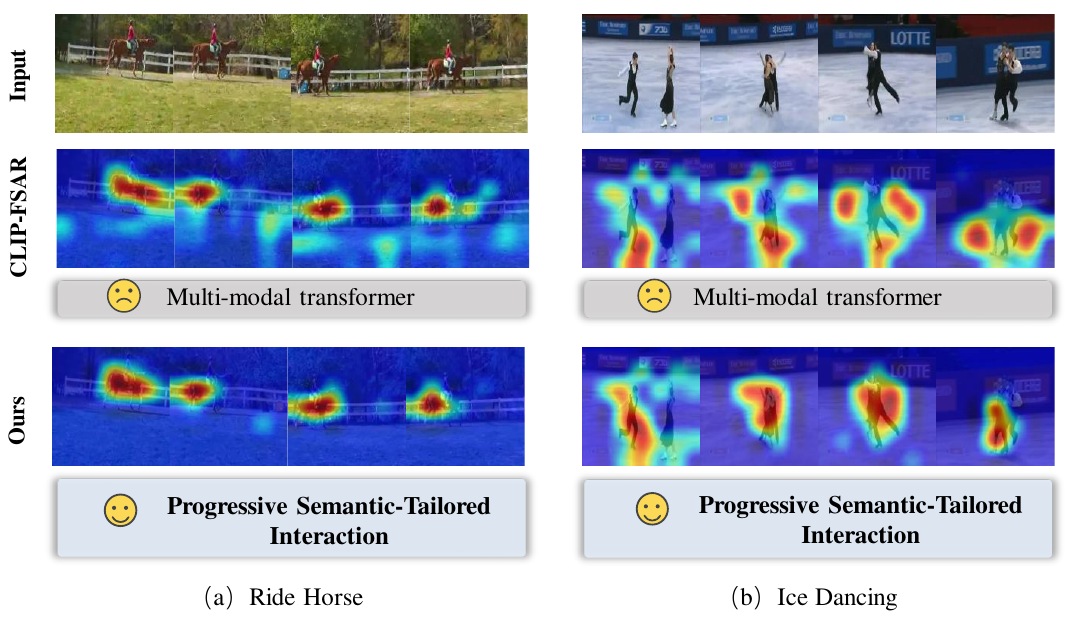}
     \setlength{\abovecaptionskip}{0.1cm}
     \caption{Attention visualization of our MVP-shot on HMDB51~\cite{kuehne2011hmdb} and UCF101~\cite{soomro2012ucf101} in the 5-way 1-shot setting. Corresponding to the original RGB images, the attention maps of the multi-modal transformer model are compared to the attention maps produced by our MVP-shot. We adopt OTAM~\cite{cao2020few} as the temporal alignment metric.}
     \label{att_map}
 \end{figure*}

\noindent \textbf{Attention$_{\!}$ Visualization$_{\!}$ of$_{\!}$ Our MVP-shot.$_{\!}$} We visualize the attention maps of our MVP-shot on HMDB51~\cite{kuehne2011hmdb} and UCF101~\cite{soomro2012ucf101} under the $5$-way $1$-shot setting (as shown in Fig.~\ref{att_map}). According to the input RGB image sequences, we compare the attention maps generated by the multi-modal transformer model using pre-trained CLIP model, as discussed in \S\ref{ablatio}, with the attention maps produced by our MVP-shot, which refers to Progressive Semantic-Tailored Interaction module (PSTI). The attention map generated with a multi-modal transformer often contains a significant number of unrelated or distracting focus areas. In contrast, the attention maps generated by our MVP-shot, which uses the PSTI module, reduce attention towards unrelated objects and the background and focus more on action-related objects. Specifically, the frames in “Ride Horse” focus on the person on the horse, and the frames in “Ice Dancing” focus on dancers on the ice. These attention Visualization maps provide empirical evidence of the efficacy of our PSTI module, which make use of semantic information provided by class names to focus on action-related areas.
 \section{Conclusion}
 \label{conclusion}
 In this work, we first point out that single-scale feature alignment in previous FSAR methods is insufficient to reflect the speed diversity of action instances. To alleviate this problem, a novel MVP-shot framework is proposed to obtain and align multi-velocity action features in a progressive way. A Progressive Semantic-Tailored Interaction module is designed to capture trustworthy multi-velocity feature representation. Subsequently, a Multi-Velocity Feature Alignment module is presented to measure the similarity of multi-velocity action features. By this means, our MVP-shot enables more robust and more accurate few-shot feature matching. Experimental results demonstrate that MVP-shot yields solid performance gains and surpasses all the state-of-the-arts on the four standard benchmarks. 
 
 However, this is the first cursory exploration of achieving multi-velocity feature matching in FSAR. Though the first step is always not elegant, more attempts and tricks in terms of multi-velocity features will be interesting in the future.

\bibliographystyle{IEEEtran}
\bibliography{ref}

\begin{IEEEbiography}[{\includegraphics[width=1in,height=1.25in,clip,keepaspectratio]{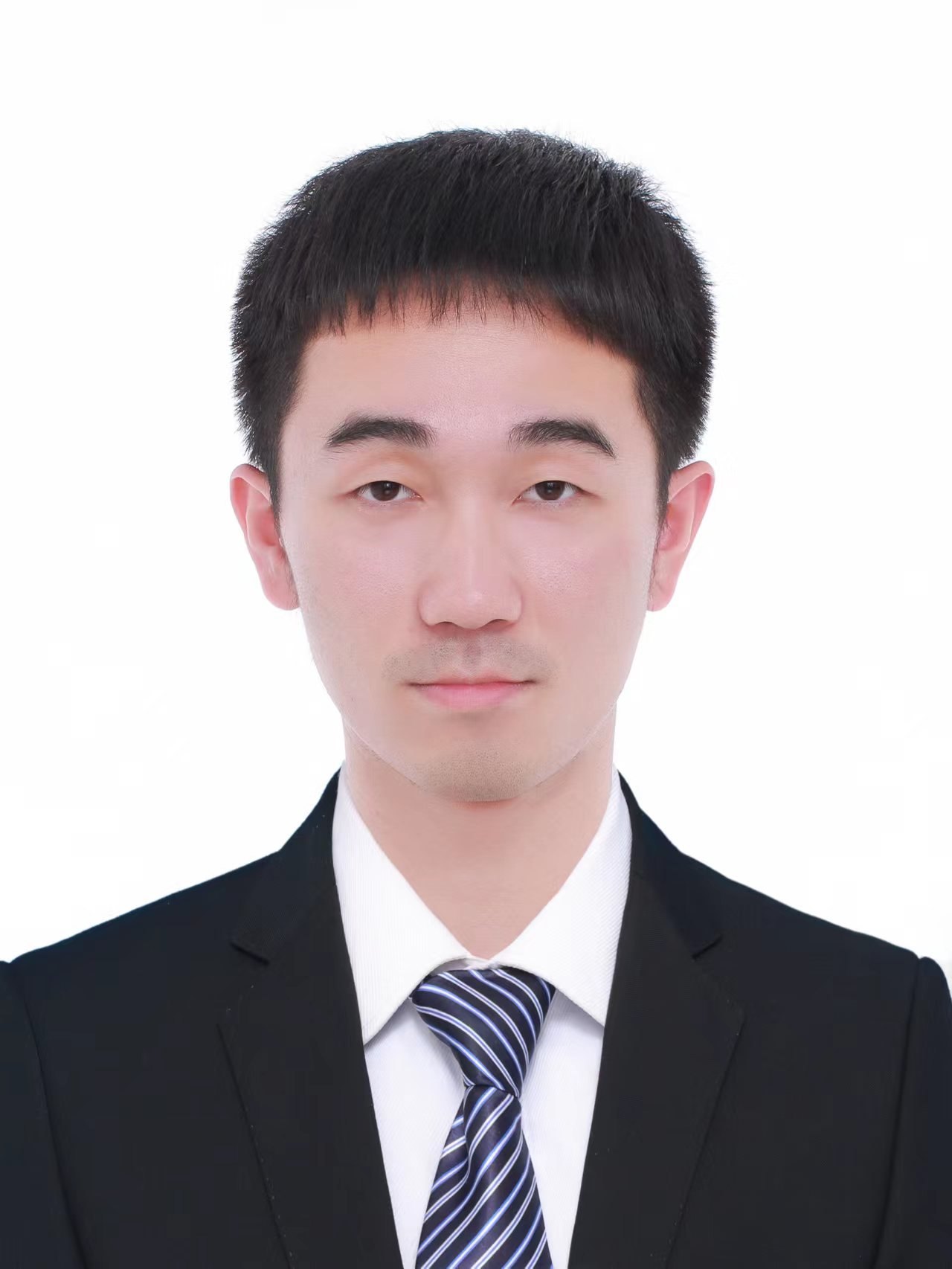}}]{Hongyu~Qu}
received the B.S. degree from Nanjing Forestry University, Nanjing, China. He is now a Master student in the School of Computer Science and Engineering at Nanjing University of Science and Technology. His research interests include video understanding and video action recognition.
\end{IEEEbiography}
\begin{IEEEbiography}[{\includegraphics[width=1in,height=1.25in,clip,keepaspectratio]{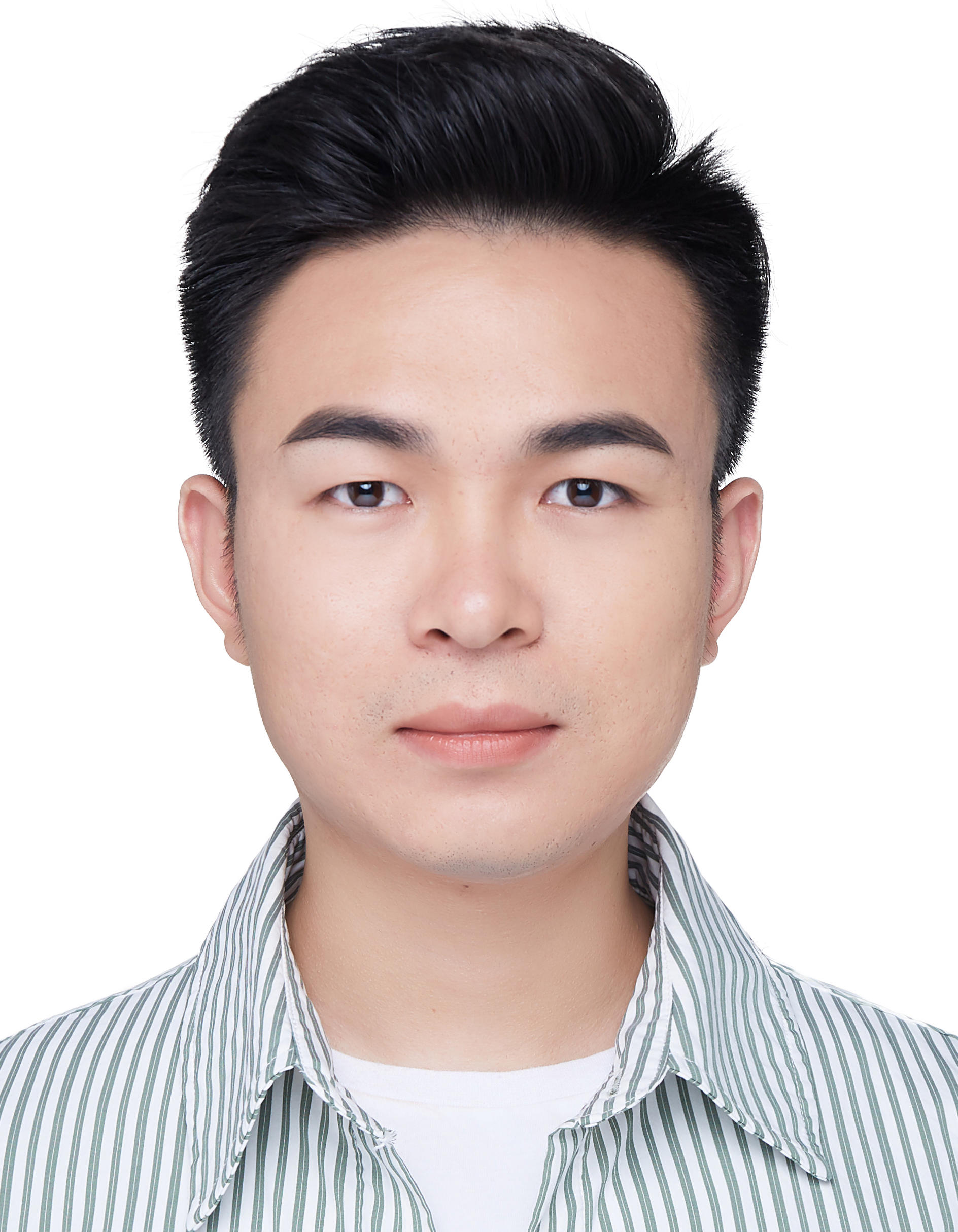}}]{Rui~Yan} 
received the Ph.D. degree at Intelligent Media Analysis Group (IMAG), Nanjing University of Science and Technology, China. He is currently an Assistant Researcher at the Department of Computer Science and Technology, Nanjing University, China. He was a research intern (part-time) at ByteDance from Jan. 2022 to Aug. 2022. He was a research intern (part-time) at Tencent from Sep. 2021 to Dec. 2021. He was a visiting researcher at the National University of Singapore (NUS) from Aug. 2021 to Aug. 2022. He was a research intern at HUAWEI NOAH'S ARK LAB from Dec. 2018 to Dec. 2019. His research mainly focuses on Complex Human Behavior Understanding and Video-Language Understanding. He has authored over 20 journal and conference papers in these areas, including IEEE TPAMI, IEEE TNNLS, IEEE TCSVT, CVPR, NeurIPS, ECCV, and ACM MM, etc.
\end{IEEEbiography}
\begin{IEEEbiography}[{\includegraphics[width=1in,height=1.25in,clip,keepaspectratio]{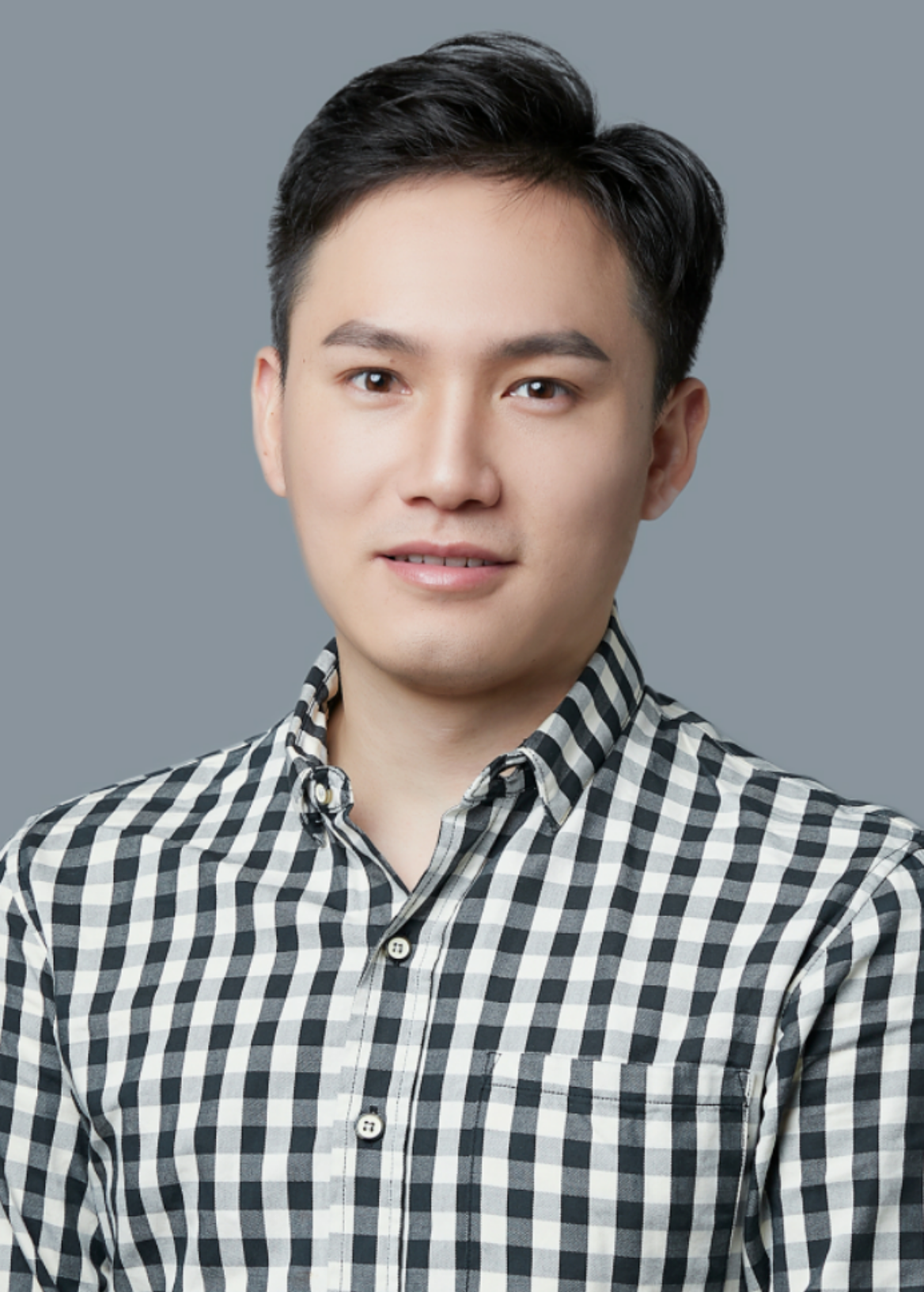}}]{Xiangbo~Shu} (Senior Member, IEEE) is currently a Professor in School of Computer Science and Engineering, Nanjing Univesity of Science and Technology, China. Before that, he also worked as a visiting scholar in National University of Singapore, Singapore. His current research interests include Computer Vision, and Multimedia. He has authored over 80 journal and conference papers in these areas, including IEEE TPAMI, IEEE TNNLS, IEEE TIP, CVPR, ICCV, ECCV, ACM MM, etc. He has received the Best Student Paper Award in MMM 2016, and the Best Paper Runner-up in ACM MM 2015. He has served as the editorial
boards of the IEEE TNNLS, and IEEE TCSVT. He is also the Member of ACM, the Senior Member of CCF, and the Senior Member of IEEE.
\end{IEEEbiography}
\begin{IEEEbiography}[{\includegraphics[width=1in,height=1.25in,clip,keepaspectratio]{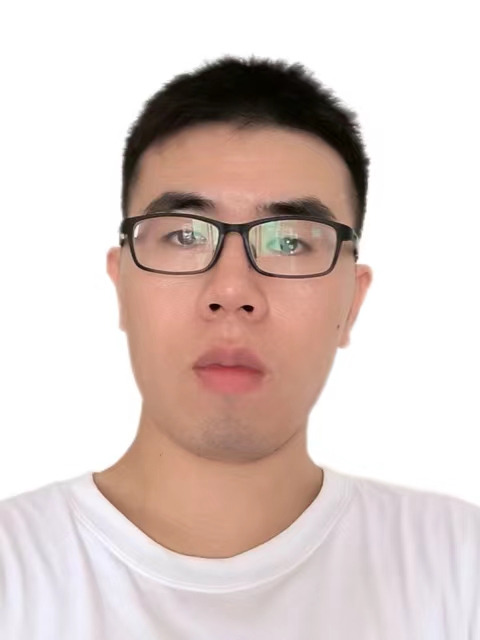}}]{Hailiang~Gao}
is currently a Master's Degree Candidate in the School of Computer Science and Engineering, Nanjing University of Science and Technology, China. Her current research interest is Computer vision and Deep Learning.
\end{IEEEbiography}
\begin{IEEEbiography}[{\includegraphics[width=1in,height=1.25in,clip,keepaspectratio]{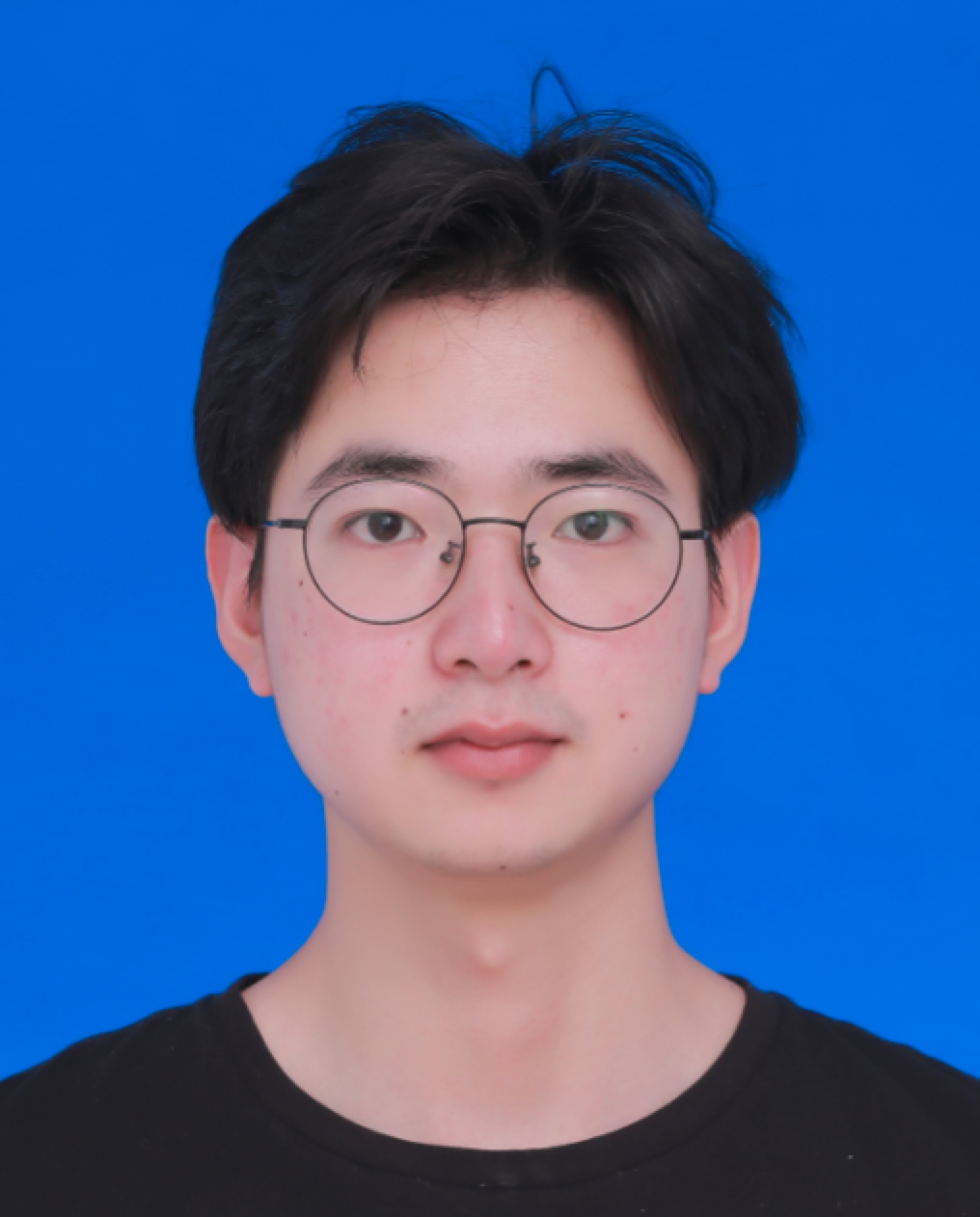}}]{Peng~Huang}
 is now a Ph.D. student in the School of Computer Science and Engineering at Nanjing University of Science and Technology. His main research interests include computer vision and multimedia analysis.
\end{IEEEbiography}
\begin{IEEEbiography}[{\includegraphics[width=1in,height=1.25in,clip,keepaspectratio]{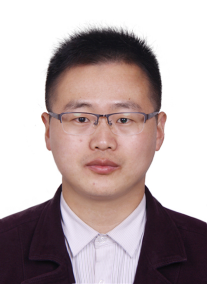}}]{Guo-Sen~Xie}
received the Ph.D. degree in pattern recognition and intelligent systems from
the National Laboratory of Pattern Recognition, Institute of Automation, Chinese Academy of
Sciences, Beijing, China, in 2016. From 2014 to 2015, he worked as a visiting scholar at the National University of Singapore, Singapore. He is a Professor at the School of Computer Science and Engineering, Nanjing University of Science and Technology, China. His research results have been expounded in more than 30 publications at prestigious journals and prominent conferences, such
as IEEE TPAMI, IJCV, IEEE TIP, IEEE TNNLS, NeurIPS, CVPR, ICCV, and ECCV. He has received the Best Student Paper Award in MMM 2016. His research interests include computer vision and machine learning.
\end{IEEEbiography}

\end{document}